# Vision-Enabled LLMs in Historical Lexicography: Digitising and Enriching Estonian-German Dictionaries from the 17th and 18th Centuries[*]

Madis Jürviste[1,2], Joonatan Jakobson[1]

**Abstract**

This article presents research conducted at the Institute of the Estonian Language between 2022 and 2025 on the application of large language models (LLMs) to the study of 17th and 18th-century Estonian dictionaries. The authors address three main areas: enriching historical dictionaries with modern word forms and meanings; using vision-enabled LLMs to perform text recognition on sources printed in Gothic script (Fraktur); and preparing for the creation of a unified, cross-source dataset. Initial experiments with J. Gutslaff's 1648 dictionary indicate that LLMs have significant potential for semi-automatic enrichment of dictionary information. When provided with sufficient context, Claude 3.7 Sonnet accurately provided meanings and modern equivalents for 81% of headword entries. In a text recognition experiment with A. T. Helle's 1732 dictionary, a zero-shot method successfully identified and structured 41% of headword entries into error-free JSON-formatted output. For digitising the Estonian-German dictionary section of A. W. Hupel's 1780 grammar, overlapping tiling of scanned image files is employed, with one LLM being used for text recognition and a second for merging the structured output. These findings demonstrate that even for minor languages LLMs have a significant potential for saving time and financial resources.

[*] This article was written with support from the Estonian national programme EKKD-III1 "Application of Large Language Models in Lexicography: New Opportunities and Challenges". The authors thank Eleri Aedmaa for technical advice and assistance in formulating queries for language models.
[**] Preprint submitted to arXiv.org.
[1] University of Tartu.
[2] Institute of the Estonian Language.

**Introduction**

Large language models, which saw significant acceleration in development following the introduction of the transformer architecture in 2017, have had a major impact on lexicography since late 2022. Initial integrations of LLMs with dictionary systems appeared just a few months after the release of ChatGPT (De Schryver & Joffe 2023) and language models are ever since receiving increasingly widespread attention at major lexicography conferences (eLex, EURALEX). However, adoption of LLMs in historical lexicography has been considerably more limited.

This article examines the application of large language models to the study of historical dictionaries at the Institute of the Estonian Language (EKI) from 2022 onward. The work presented, which also includes a brief discussion of future plans, is drawn primarily from the authors' own research areas.

Using an empirical approach, we address two primary questions.

1. How can we make historical sources accessible to language researches and the general public when the language in those sources has changed so significantly that it is difficult to understand for modern readers?
2. How can we leverage language models for text recognition and structured parsing of those dictionaries, which were published in Fraktur centuries ago? Answering these questions will lay the groundwork for presenting historical dictionary content in a more accessible and comprehensive format.

**1. Timeline: The Use of Large Language Models in Historical Lexicography at the Institute of the Estonian Language**

The widespread adoption of large language models in late 2022 is the culmination of decades of computational approaches to language understanding. Much of early computational language analysis in the 1990s focused on statistical patterns, including frequency-based approaches that counted how often words occurred together in large text collections. These statistical models could predict probable word sequences based on observed frequency — for instance, learning that "historical" usually precedes "linguistics" — but they were limited to immediate context (typically 2-3 surrounding words) due to computational constraints. Each additional word included in context exponentially increased the required calculations, making broader analysis impractical. (Wang et al. 2024)

A significant advance arrived around 2013 with neural network approaches, exemplified by Word2Vec (Mikolov et al. 2013), which could efficiently capture semantic relationships between words. These systems learned that words appearing in similar contexts typically had related meanings—positioning words like "king" and "monarch" close together in their corresponding vector representations. However, each word was treated as fixed entity, without consideration to semantic diversity. As these models could not differentiate between languages or contextual variations, the Estonian word *sure* (imperative form of *die*) would be computationally indistinguishable from English *sure* ('certain').

The major breakthrough came in 2017 when Google researchers introduced the transformer architecture (Vaswani et al. 2017). This enabled models to dynamically consider their surrounding context—recognizing that identical word forms could carry entirely different meanings depending on their linguistic environment. This innovation led to powerful pre-trained models that could be fine-tuned for specific tasks such as translation or summarisation, and by 2022, these models had evolved into large language models such as ChatGPT, capable of performing diverse language tasks without requiring task-specific training. The implications of this technology to lexicography became apparent immediately follwing the public release of ChatGPT in November 2022 (De Schryver & Joffe 2023). At EKI, the initial application of language models in the context of historical lexicography was for practical tasks such as composing simple Python scripts to analyse datasets compiled manually.

Since 2024, a dedicated working group (Madis Jürviste, Tiina Paet, Sven-Erik Soosaar, assisted since 2025 by Joonatan Jakobson) has been investigating the effectiveness of language models in analysing historical dictionary content (Jürviste et al. 2025a), as well as the application of LLMs for optical character recognition (OCR) and structured parsing of such documents. For the general public, EKI has developed Lummaja, a conversational agent based on RAG architecture. This tool enables the visitors of the Estonian National Museum, regardless of their linguistic background, to explore the contents of historical dictionaries and learn about their authors. For researchers investigating the history of specific, a unified query system named Wilhelm is in development. It is designed to provide language researchers with a tool to retrive initial information from numerous valuable sources, including dictionaries and corpora compiled at EKI, the University of Tartu's old written language corpus (VAKK), and other resources. While Lummaja functions as a conversational agent, Wilhelm provides concise overviews of dictionary content with precise source references.

Initially, LLMs could not process image files directly. Instead, models from OpenAI, Anthropic, and Google relied on external Python libraries for analysing PDFs, resulting the OCR output being unintelligible for non-native PDFs (i.e. files converted not from a text document, but scanned as images). Thus, we started our research by manually inputting short sample datasets and assessing whether large language models were capable of analysing Estonian vocabulary from the 17th and 18th centuries. These initial experiments[3] proved successful (see Jürviste et al. 2025a). Then a promising development emerged in August 2024 when Anthropic released Claude 3.5 Sonnet, which was among the first truly vision-capable language models. Unlike its predecessors, which utilized traditional Python libraries for retrieving information from scanned documents, Claude 3.5 could read image file content from base64-encoded data. In essence, the language model could extract text directly from an image file (e.g., a PNG) without requiring external preprocessing steps.

This new capability enabled the working group to outline specific experiments for digitising historical dictionaries. The initial test case was the dictionary section of A. T. Helle's 1732 grammar (Helle 1732, hereafter "Vocabularium"), as this source, despite its importance to the history of the Estonian language, had not yet been converted into a machine-readable and editable format. The subsequent project will focus on the Estonian-German dictionary section (hereafter "Wörterbuch") of A. W. Hupel's 1780 grammar. The digitisation process for both Helle's and Hupel's dictionaries is described in detail in Chapter 3.

---

[3] Carried out by Madis Jürviste, Tiina Paet and Sven-Erik Soosaar.

In addition to digitisation, the working group has also investigated the feasibility of using LLMs to enrich historical dictionaries with modern word forms and meanings. Initial experiments in this area, involving the semi-automatic supplementation of the dictionary section (hereafter "Nomenclator") of Johannes Gutslaff's 1648 grammar, are detailed in Chapter 2. These technological advances, combined with the preparatory work described here, pave the way for a unified database of historical dictionaries and enables the creation of a comprehensive "Dictionary of Dictionaries" based on this cross-source data. Future work will also involve extending these experiments to more recent lexicographical sources, starting with F. J. Wiedemann's dictionary (first edition 1869).

The following table gives a concise overview of these dictionaries:

| Year | 1637 | 1648 | 1660 | 17XX | 1732 | 1780 |
|---|---|---|---|---|---|---|
| Author | H. Stahl | J. Gutslaff | H. Göseken | S. H. Vestring | A. T. Helle | A. W. Hupel |
| Short. title | "Vocabula" | "Nomenclator" | "Farrago" | "Lexicon" | "Vocabularium" | "Wörterbuch" |
| Ling. area* | eN | eS | eN | eN | eN | eN, eS |
| Entries | 2309 | 1859+ | 9000+** | 9900** | 7000** | 17.000** |
| Publication (print / manuscr.) | P 1637 P 1974$^\partial$ | P 1648 P 2002$^\partial$ | P 1660 P 2010 | M ~1720$^\alpha$ P 1998 | P 1732 P 2006$^\partial$ | P 1780 |
| Previous research | Kikas 2002 | Lepajõe 1998; Viitkar 2005 | Kingisepp et al. 2010 | Leesmann 1993 | Kilgi, Ross (*in* Helle 2006) | Jürjo 2004 |
| Direction | DE-ET | DE-ET | DE-ET | ET-DE | ET-DE | ET-DE |

\* eN, eS: *resp.* Northern Estonian, Southern Estonian.
\*\* Approximations.
$^\alpha$ Manuscript, published in print in 1998.
$^\partial$ Facsimile.

Table 1. Overview of the Estonian dictionaries from the 17th and 18th centuries (1637–1780) discussed in the present article.

## 2. Enriching Historical Dictionaries with Modern Word Forms and Meanings

Historical dictionaries are rich sources of information about the language, worldviews, society, and culture of their time. However, due to significant linguistic changes, this material is often opaque to contemporary dictionary users. To address this, the modern editions of two foundational historical dictionaries – Heinrich Stahl's 1637 "Vocabula" and Göseken's 1660 "Farrago" (Kikas 2002; Kingisepp et al. 2010) – have reversed the original German-Estonian translation direction, allowing headwords to be browsed using Estonian as the source language. Furthermore, the headwords have been supplemented with modernised word forms and meanings.

Similarly, Salomo Heinrich Vestring's "Lexicon Esthonico Germanicum", compiled around 1720, has also been published in print (Vestring 1998 [17XX]), although it lacks modern word definitions. The dictionaries of Gutslaff (1648), Helle (1732) and Hupel (1780) were previously available only as scanned PDFs, though Viitkar (2005) and Lepajõe (1998) have made portions of this material more accessible. The following examples illustrate the challenges these dictionaries pose for contemporary readers:

1. Gutslaff (1648: *Kranßaug – sillipuu*. Neither the German headword *Kranßaug* nor the Estonian equivalent *sillipuu* appear in contemporary standard dictionaries of either language. (According to (Viitkar 2005: 138), *sillipuu* means 'foxtail grass (plant)', while Kingisepp et al. give the meaning 'strychnine tree' for the same entry in Göseken (1660: 265).)
2. Vestring (1998 [17XX]: 45): *Jätksem – Comparat. Das mehr verschlagt, Verschlägsam*. (likely meaning 'that which suffices for more, for longer'; Jürviste 2025: 46). Despite Vestring's inclusion of a usage example, the meaning of this word remains opaque to contemporary readers relying solely on this dictionary.
3. Helle (1732: 102): *kaarlad – Schelbeere*. Here, the German equivalent *Schelbeere* is obsolete, and while the Estonian word *kaarlad* ('cloudberries') appears in Estonian dialects (EMS, s. v. kaarlad), users would need to consult specific Estonian dialect dictionaries to understand the meaning of this entry.

The new editions of Stahl (Kikas 2002) and Göseken (Kingisepp et al. 2010) described above were manually compiled, a process that required significant time and expertise. However, given LLMs' demonstrated success with German language processing (as discussed in Jürviste et al. 2025a), it is worth investigating whether the remaining dictionaries (Vestring, Helle, Hupel) could be semi-automatically enriched with: (a) modern equivalents for historical headwords and (b) corresponding contemporary definitions.

To evaluate whether semi-automatic enrichment is viable, we selected Gutslaff's "Nomenclator" for our initial experiments. This source was chosen for two primary reasons. It has already been the subject of previous research (Lepajõe 1998; Viitkar 2005), and its scope is well-suited for the experiment, with the surviving incomplete copies containing 1859 headwords, according to the manual transcription completed at EKI in 2024–2025.

Given that all the dictionaries under consideration – whether German-Estonian or Estonian-German – were compiled within the same linguistic area[4] and also in the same cultural sphere (all authors were Baltic German Lutheran ministers) (see Jürviste 2023; Jürviste 2025), we deemed it expedient to conduct an initial mapping of headwords using language models. The goal was to determine which of Gutslaff's headwords also appear in the other five dictionaries, and which of those sources contain contextual information that could be used for enriching Gutslaff's entries. In essence, this process involved providing the language model with existing dictionary data to simulate the manual workflow of a lexicographer.

We employed a two-stage process for this task: OpenAI's GPT-4o (OpenAI 2024) for initial mapping (via its chat application) and Anthropic's Claude 3.7 Sonnet (via API) for reviewing GPT-4o's results. GPT-4o was provided with four dictionary files in CSV format: Gutslaff's "Nomenclator" (the version transcribed at EKI) served as the basis for headword ordering, while the modern editions of Stahl's "Vocabula", Göseken's "Farrago" and Vestring's "Lexicon" were used as the source texts for supplementary information. The model was instructed to process each headword from Gutslaff's list and search the other sources for corresponding entries or related information. The initial table generated by GPT-4o was subsequently reviewed line-by-line by Claude, which supplemented the output with its own findings and commentary. The result was a table in which most headwords from Gutslaff were matched with multiple corresponding entries from the other dictionaries in both Estonian and German.

---

[4] The linguistic landscape of 17th and 18th-century Estonia was characterised by significant differences between the Northern and Southern dialects and the absence of a unified standard language. This dialectal split is reflected in the lexicographical works of the period: Stahl, Göseken, Vestring, and Helle focused on Northern Estonian vocabulary, whereas Gutslaff catalogued Southern Estonian words. Hupel, in contrast, adopted a more nuanced methodology by including numerous regional markers.

| Gutslaff's Estonian headword | Ubbene 'apple' |
|---|---|
| Gutslaff's German equivalent (EE-DE direction) | Apffel |
| Stahl's Estonian headword | Aun |
| Stahl's modernised Estonian equivalent | õun |
| Stahl's German equivalent (EE-DE direction) | Apffel |
| Göseken's German equivalent (EE-DE direction) | apffel |
| Göseken's Estonian word form (EE-DE direction) | Oun |
| Göseken's Estonian headword | õun |
| Vestring's Estonian headword | Oun |
| Vestring's German equivalent | Der Apffel |
| Vestring's example sentence in Estonian | Ouna Südda. |
| Vestring's Estonian example sentence translation into German | Das Korn gehäuse im Apffel |

Table 2. Example of dictionary information mapping: Information added by GPT-4o and Claude 3.7 for Gutslaff's Estonian headword (from manually edited table)

However, the language models often failed to create precise matches due to several factors, including homonyms, highly variable orthography of Estonian words and the semantic ambiguity of German equivalents. Consequently it was necessary to manually clean a portion of this aggregated table to remove obviously erroneous content [5]. The resulting table provided reliable contextual information for enriching Gutslaff's dictionary entries.

The initial test was conducted on a subset of 342 entries. The language model (Anthropic's Claude 3.7 Sonnet, API query on June 3, 2025) examined Gutslaff's words row by row and added a modern equivalent, definition, and, where necessary, additional commentary. The results were promising: the model successfully provided correct modern forms and meanings for 81% of the 342 entries. Of the remainder, 11% required minor editing, while 8% of entries needed complete manual revision.

| Old EE | Modern EE | Old DE | Modern DE | Comment |
|---|---|---|---|---|
| Auw | au | Ehre | Ehre | Auw means honour/esteem. German Ehre has not changed. |
| auwus | aus | ehrbar | ehrbar | auwus is modern aus (honourable). German ehrbar has not changed. |
| Bicken | kauss | Becken | Becken | According to this context, Bicken means bowl. German Becken has not changed. |

---

[5] Sven-Erik Soosaar carried out most of the manual work during this research stage.

| | | | | |
|---|---|---|---|---|
| Bôdi | pood | Bude | Bude | According to this context, Bôdi means shop. German Bude has not changed. |
| Carman | kott | Beutel | Beutel | According to this context, Carman means bag. German Beutel has not changed. |
| dunckel | tume | Brand | Brand | According to this context, related to fire/burning Brand. German has not changed. |
| effardama | ähvardama | Dräuwen | drohen | effardama is modern ähvardama. German Dräuwen is today drohen. |

Table 3. Example of enriching historical dictionary (Gutslaff 1648) headword entries with Estonian and German modern forms (content of "Comment" column is from Claude)

While this experiment warrants replication with a larger dataset and a better refined prompt (i.e., for a more precise definition of *contemporary headword*), the results clearly illustrate that an LLM's output can substantially assist in the manual enrichment of dictionary content when provided with sufficient context. The primary benefit is the time saved for a human editor. The language model can pre-fill a substantial portion of the dictionary – specifically, the straightforward entries that pose no interpretive challenges. In this way, the editor can focus their efforts on more problematic entries requiring nuanced analysis. For instance, the Estonian equivalent *dunckel* (see table above) necessitates manual investigation to ascertain whether it represents a misplaced German word or the author's proposed historical South Estonian word form.

## 3. Text Recognition of Historical Dictionaries Using LLMs

In August 2024, Anthropic made its Claude 3.5 language model available to the general public. A key update in this version is the ability to analyze image data directly from its base64 encoding, whereas previous versions relied on external OCR libraries (e.g., PyPDF). This means the model can now effectively "read" scanned documents by identifying visual linguistic patterns, such as Fraktur script, directly from the image's raw data.

Since language models can both recognize text and structure data, they can be instructed to output the scanned text in a structured format like JSON, CSV or XML. Initial experiments with Claude's chat application (Nov 22, 2024) confirmed this capability, showing that it could successfully generate structured output from the 18th-century dictionary pages of August Wilhelm Hupel (1780) and Anton Thor Helle (1732). The assuring results of these preliminary experiments, which yielded only minor errors, motivated the working group to attempt a full-scale OCR of both aforementioned dictionaries. The two primary objectives were to (1) test the model's vision capabilities on a larger dataset, and to (2) produce a fully machine-readable and machine-editable versions of these dictionaries. A significant long-term outcome of this work is the potential to compile the first modern editions of these historical sources, a task not previously undertaken.

## 3.1 OCR: Anton Thor Helle's 1732 "Vocabularium"

We began by testing Anthropic's **Claude 3.5 Sonnet** to perform OCR and generate structured output with the dictionary section of A. T. Helle's 1732 grammar (Helle 1732). This source was well-suited for a pilot study due to the high quality of its digital scans (Estonian National Library's digital archive DIGAR version: Helle 1732), and its manageable scope of 129 pages, which was optimal for an initial dictionary-wide experiment.

The novelty of this experiment[6] lies in its use of a zero-shot method[7] in applying LLMs for text recognition. Traditionally, this task has required specialised software (e.g., Kraken, Transkribus, eScriptorium) for historical printed and manuscript sources, and while these tools are effective, they demand substantial preliminary work in order to achieve reasonable error rates. In this field, a Character Error Rate (CER) below 5% is generally considered a successful outcome.[8]

Traditional OCR software requires two main preparatory steps. First, the source material must be segmented. This process, which ranges from manual to semi-automatic, identifies the precise location of text within the image (e.g., by marking text blocks). Second, a *ground truth* dataset must be created, which consists of a manually transcribed, error-free sample of the text. The quality and volume of this ground truth determines the accuracy of the final output, with more extensive preparation leading to a lower error rate (a CER well below 5%).

While such extensive preparation is justifiable for large and uniform copies or manuscripts, it would be rather impractical for Helle's "Vocabularium", as the effort required would amount to simply transcribing the entirety of this volume manually. Crucially, LLMs do not require this *ground truth* preparation: our initial tests already produced a near-perfect output in a structured CSV format. Furthermore, a fundamental problem with Transkribus-like solutions is its unstructured output: even if the result is of very high quality (characterised by a low error rate), it is not structured. This "pure OCR" might be entirely sufficient for prose, but ill-suited for a dictionary, where each entry is comprised of distinct data fields (e.g., headword, definition, example). We therefore decided to forgo traditional software and rely solely on a *zero-shot* LLM approach.

Our first obstacle was calibrating the API query parameters to replicate the high-quality results we initially observed in the Claude 3.5 chat interface. LLMs are controlled by several settings, most notably temperature, which governs the creativity of the output. On a scale from 0.0 to 1.0, a lower the temperature produces deterministic, focused responses, while a high temperate allows for wider variation and unpredictability, resembling creativity. These parameters must be tuned manually when accessing the language model via its API. Consequently, our initial API queries produced lower-quality output compared to the promising results from the chat application. However, the crucial advantage of using the API is reproducibility; by fixing the parameters, we achieved more consistent and uniform results across all runs, which was essential for the experiment.

A second, more significant obstacle was page segmentation. The chat application handled the dictionary's two-column format perfectly, but the API failed to properly interpret the layout.

---

[6] Carried out by Madis Jürviste, Joonatan Jakobson, Tiina Paet and Sven-Erik Soosaar.
[7] In a *zero-shot* approach there is no comprehensive example data given to the language model.
[8] CER is a standard metric used to measure text recognition accuracy by calculating the percentage of incorrect characters (substitutions, insertions, deletions) compared to the reference text, using Levenshtein distance.

This resulted in severe errors in the output: large portions of the page were omitted (including the entire right column), and the model often read entries horizontally across the page instead of vertically. Faced with these issues, we considered manual segmentation, but this option was dismissed as it would have been prohibitively time-consuming.

The release of a new model, **Claude 3.7 Sonnet** (February 24, 2025), resolved both of our previous obstacles. This updated version successfully handled page segmentation for the two-column layout directly from PNG files. Furthermore, the quality of output via its API improved drastically, with a substantially lower error rate in comparison to its predecessor, Claude 3.5. This breakthrough allowed us to continue the experiment using the consistent and reproducible API environment. To facilitate this large-scale task, we developed (with the assistance of Claude) a web application, **Anthonius**, which helped us manage the submission of the 129 dictionary pages and to finally export the resulting JSON output as a unified CSV file for analysis.

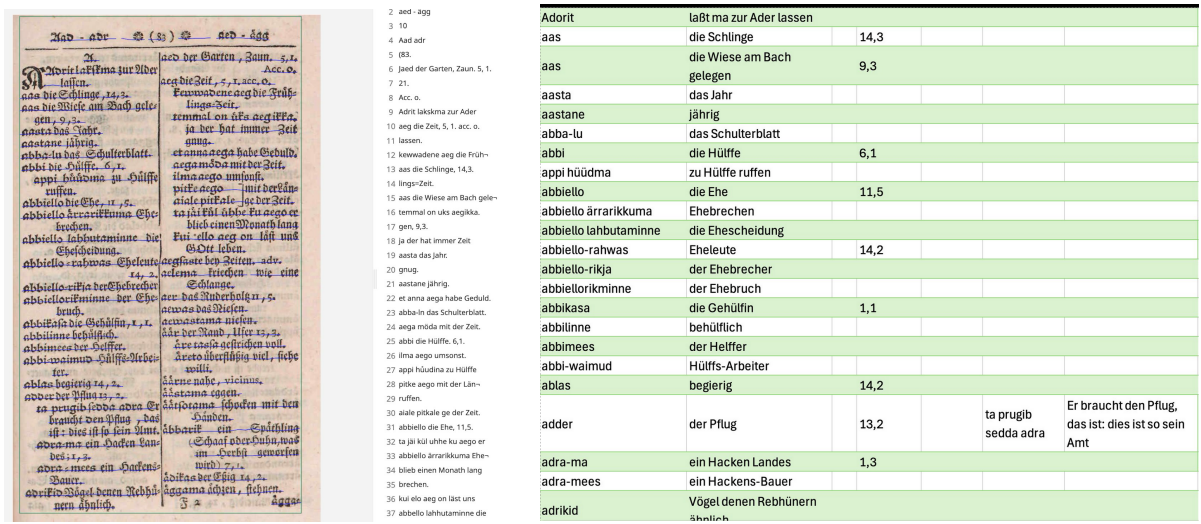

Figure 1. Comparison of dictionary digitization outputs. Left: Raw text transcription for Helle's "Vocabularium" from Transkribus. Right: Structured, tabular output from Claude 3.7 Sonnet in a zero-shot setting.

Our prompt for Helle's "Vocabularium" was divided into three main sections. First, we instructed the model to preserve the dictionary's printed structure, directing it to read vertically down each column and to retain all text, including partial entries at page breaks. Second, we emphasized the need for accuracy, requiring the model to replicate original orthography, punctuation, and formatting, with specific rules for the presentation of grammatical information (such as case and conjugation ). Finally, we defined an output schema for constructing each entry with nine corresponding fields: Estonian headword, Estonian synonyms, German translation equivalent, German synonyms, Latin explanation, part of speech, grammatical information, and both Estonian and German multi-part linguistic elements containing whitespaces.

To analyze the results, Claude's JSON output was first converted into CSV format, which allowed for a systematic review of the dictionary content in a spreadsheet. We then evaluated the experiment's success by comparing the LLM's raw output against a manually edited, corrected version[9]. This evaluation considered several traditional OCR quality metrics, of which CER proved to be the most informative for assesing the model's accuracy.

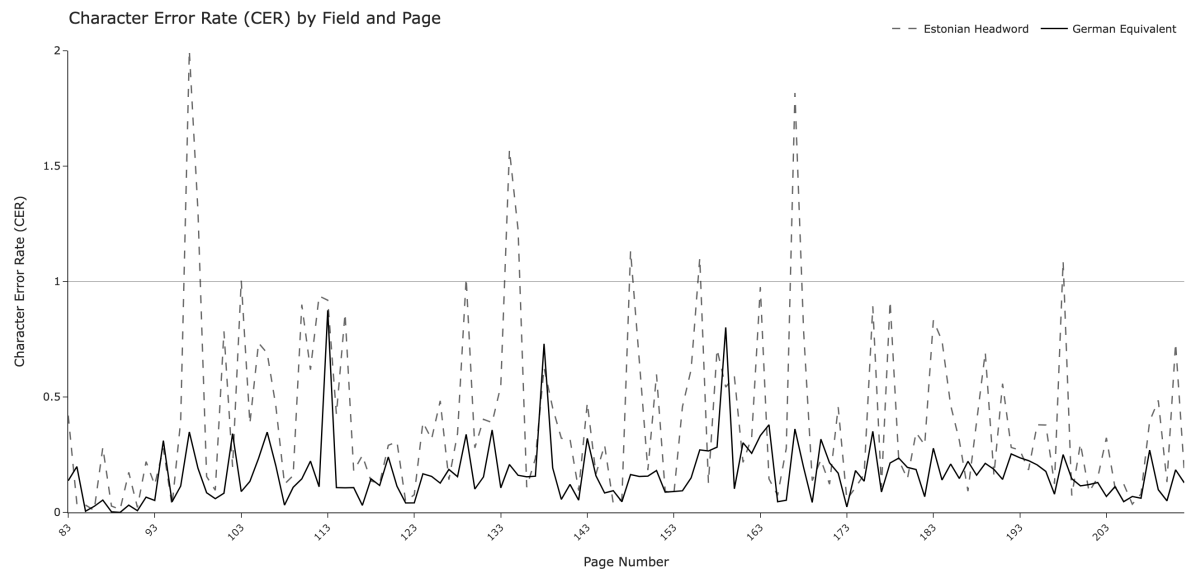

Figure 2. Character Error Rate (CER) by page for Helle's dictionary, comparing Estonian Headwords (dashed line) to the German Equivalents (solid line).

The dramatic fluctuations in the Character Error Rate (CER), clearly visible in the graph, are a direct result of the model's page-by-page processing method. Because Claude processed each dictionary page as an independent task, it had no memory of its previous output. Consequently, an error made at the top of the page, such as misinterpreting a single character (e.g., reading *õ* as *ö* or *f* for *s*), was often propagated throughout the rest of that page. This tendency is the primary reason for extreme CER spikes seen on certain pages, which was particularly pronounced for the Estonian Headword field (dashed line).

Quantitatively, the model's performance on German headwords was significantly better than on Estonian. The average **CER for Estonian headwords was 42%,** with the rate on some pages exceeding 100%. This was often due to structural errors where correctly recognized words were misplaced in the output, resulting in a complete mismatch. In contrast, the average **CER for German headwords was 17%**. We hypothesize this discrepancy is due to the greater prevalence of German over Estonian in the model's training data. Despite the high average error rates, **41% of all entries were generated perfectly**, requiring no manual corrections. A typical error in the otherwise accurate German output was the model's tendency to modernize historical spellings rather than preserving the original orthography.

The model's errors can be grouped into several categories. First, it exhibited classical character recognition errors such as ***f*** for ***s*** (*ſ* in Fraktur), ***b*** for ***h*** (*lahbutaminne* for *lahhutaminne* 'divorce', *ebbitaminne* for *ehhitaminne* 'construction; adorning', *mebbe* for *mehhe* 'man's (Gen.Sg.)'), ***t*** for ***k*** (*autlik* for *auklik* 'holey, containing holes'). Second, we encountered amalgams where the LLM would merge the content from two different lines, for example, producing *wopson* from *worm* 'printing forme' (the preceding headword was *wopsima*), p. 212.

---

[9] Madis Jürviste, Sven-Erik Soosaar, Joonatan Jakobson, Tiina Paet.

The third and most peristent tendency was modernisation, where the model updated the historical forms despite our explicit instructions: *õ* for *ö*, for example, *kõrts* for *körts* 'tavern' (which was one of the most frequent attempts of modernisation), *wälk* for *walk* 'lightning' (apparently under the influence of German *der Blitz*, p. 203), there were also verb substitutions (*ausrennen* for *auftrennen* 'unravel', p. 87) and the use of modern forms instead of historical ones (*päbstlich* for *papistisch* 'papal', p. 151; *tubakat* for *tubbakat* 'tobacco (Sg.Part.)', p. 206).

Nonetheless, Claude's output was occasionally exceptional, even correcting printing errors in the original source text, such as identifying *höble-keed* as *höbbe-keed* 'silver chain', p. 110. However, as previously mentioned, the CER calculation was often skewed by structural errors, which increased the error rate even when the underlying text recognition was performed correctly.

Although the results are far from the standard error rate of <5%, the output was assuredly suitable for manual post-editing, a process that was significantly faster than transcribing the entire dataset from scratch. Perhaps more importantly, the experiment provided crucial lessons for future work. The main takeaway was that we had underestimated the structural complexity of the "Vocabularium". Our initial output schema was not granual enough to handle the frequent occurence of reference entries, synonyms and multi-word units. While constructing the prompt, we did not anticipate the prevalence of such elements. Though this data could be inferred and restructured *post-hoc* with the aid of an LLM, it is preferable to develop a more granular output schema from the outset.

Finally, this experiment called into question the suitability of CER as a primary metric for assessing the quality in this task. CER is based on Levenshtein distance and is optimized for measuring character-level errors in linear text, but its application to structured, multi-field output is problematic, as OCR errors and segmentation errors cannot be properly distinguished.

Therefore, despite the high CER, we assess the Helle "Vocabularium" experiment as a successful proof of concept. Obtaining the initial output of dictionary content took, on average, less than a minute per page, and the cost of API queries for digitising the entirety of the dictionary was approximately 10 euros. In other words, this experiment demonstrated that an LLM-based approach can be highly time-efficient and affordable as method for digitising and structuring historical dictionaries.

**3.2 OCR: August Wilhelm Hupel's 1780 "Wörterbuch"**

Following our work on Anton Thor Helle's 1732 "Vocabularium", we continued experimenting with LLM-based digitisation, applying it to the Estonian-German dictionary section ("Wörterbuch") of August Wilhelm Hupel's 1780 grammar. This text, though vital from the perspective of Estonian history of lexicography, has until now been available only as a scanned PDF, with a low-quality OCR[10] layer insufficient for direct conversion into a machine-readable format.

Our methodology for the "Wörterbuch" diverges from that employed for Helle's dictionary. We began by substantially enhancing our API workflow and prompts, and with the recent availability of vision capabilities in OpenAI's GPT models and Google's Gemini—in addition to Anthropic's Claude—we conducted an analysis to determine which model could most effectively generate structured output from an image file. We selected TEI Lex-0 XML as the output format, as it is specifically designed for the semantic markup of dictionary information and allows for a highly granular representation of the data.

---

[10] Probably using ABBYY software: https://www.riha.ee/api/v1/systems/digar/files/0ddf9de6-1997-44f1-acca-f304c97b6909 (11.09.2025).

The complexity of the TEI Lex-0 XML format nevertheless required us to manually compile a sample file, as the language models struggled to consistently generate a correct and uniform XML structure on their own. This sample, based on a single page (p. 149) containing 86 headwords, served as our ground truth for comparing the different models and processing methods.

In the first experiment, we compared models from Anthropic, Google and OpenAI. Each model was provided with identical instructions and an image of a dictionary page. The results were evaluated based on two key criteria: structural similarity (the correctness of the generated XML markup) and content similarity (the accuracy of the text recognition from the image).[11]

The experiment revealed that Anthropic's Claude and Google's Gemini models substantially outperformed OpenAI's in both structure and content recognition (with average structural similarity scores of **0.457** and **0.431**, respectively, compared to OpenAI score below **0.150**). The OpenAI models often produced incomplete results, recognising an average of only ~18 headwords out of 86, or refused to respond, citing the image's excessive complexity as justification: "I'm sorry – the image is too detailed and contains too much information for me to produce a reliable, fully-proof-read TEI-Lex0 transcription within this single response" (OpenAI o3). Due to this poor performance and higher cost, OpenAI models were excluded from further analysis.

For Anton Thor Helle's dictionary, we used the non-reasoning variant of Claude 3.7 Sonnet[12], as it yielded better results in preliminary testing. We hypothesized that reasoning models tend to over-analyse images, as their step-by-step processing, designed for complex textual tasks, may be suboptimal for direct visual recognition. To test this hypothesis, we designed an experiment comparing the reasoning and non-reasoning variants of the language models. The reasoning models performed on average 20.4% better in structure recognition, suggesting that this step-by-step reasoning process helps the model to adhere more closely to the XML schema. However, the impact on content accuracy was negligible, with only a 2.9% improvement. It should be noted, that the utility of this "thinking" is highly model-specific; for instance, with the Claude 4 Opus model, structural accuracy improved by 13%, while content accuracy actually decreased by 15.4%. Therefore, increasing the "thinking" budget is not a universal solution but a strategy whose effectiveness depends of the specific model and task.

In our experiments to date, Gemini 2.5 Pro has yielded the best results (structural accuracy 0.57 and content 0.53). However, numerous errors still persist both in the structure and the transcription (e.g. *ette **jäädma*** for *ette **säädma*** 'to put forth'). Given the high volume and density of information per page (on average, 7.1k input tokens and 17.5k output tokens), we sought to make the task more manageable for the language model by dividing the analysis of the page into smaller chunks.

---

[11] Since we wanted to compare structure and text with a unified metric, CER was not suitable, so we used the SequenceMatcher function based on the Ratcliff-Obershelp algorithm in Python's difflib library, whose output is a ratio between 0.0 (no similarity) and 1.0 (identical), taking sequence order into account in calculating overlap. The sequence created for structure comparison consisted of XML tags and their attributes; text accuracy was measured purely from the content of tags in XML: each character is a separate member in the sequence.

[12] Claude 3.7 Sonnet system card: https://www.anthropic.com/claude-3-7-sonnet-system-card (11.09.2025).

Decomposing tasks into smaller components is a common technique in OCR. For analysing documents with complex and unpredictable layouts, a segmentation stage is typically used to identify and separate different elements – such as text, images, and tables – for individual processing (see e.g. Zhang et al. 2025). Given the uniform layout of the dictionary, we were able to use a simpler overlapping tiling approach without needing precise segment detection. As the dictionary is laid out in two columns, we first divided each page in half using traditional image processing libraries (PIL – Python Image Library). These two columns were then used to create four smaller, overlapping segments. This overlap, where each segment included the content from the adjacent segments, was designed to mitigate the risk of omitting text due to inexact visual partitioning.

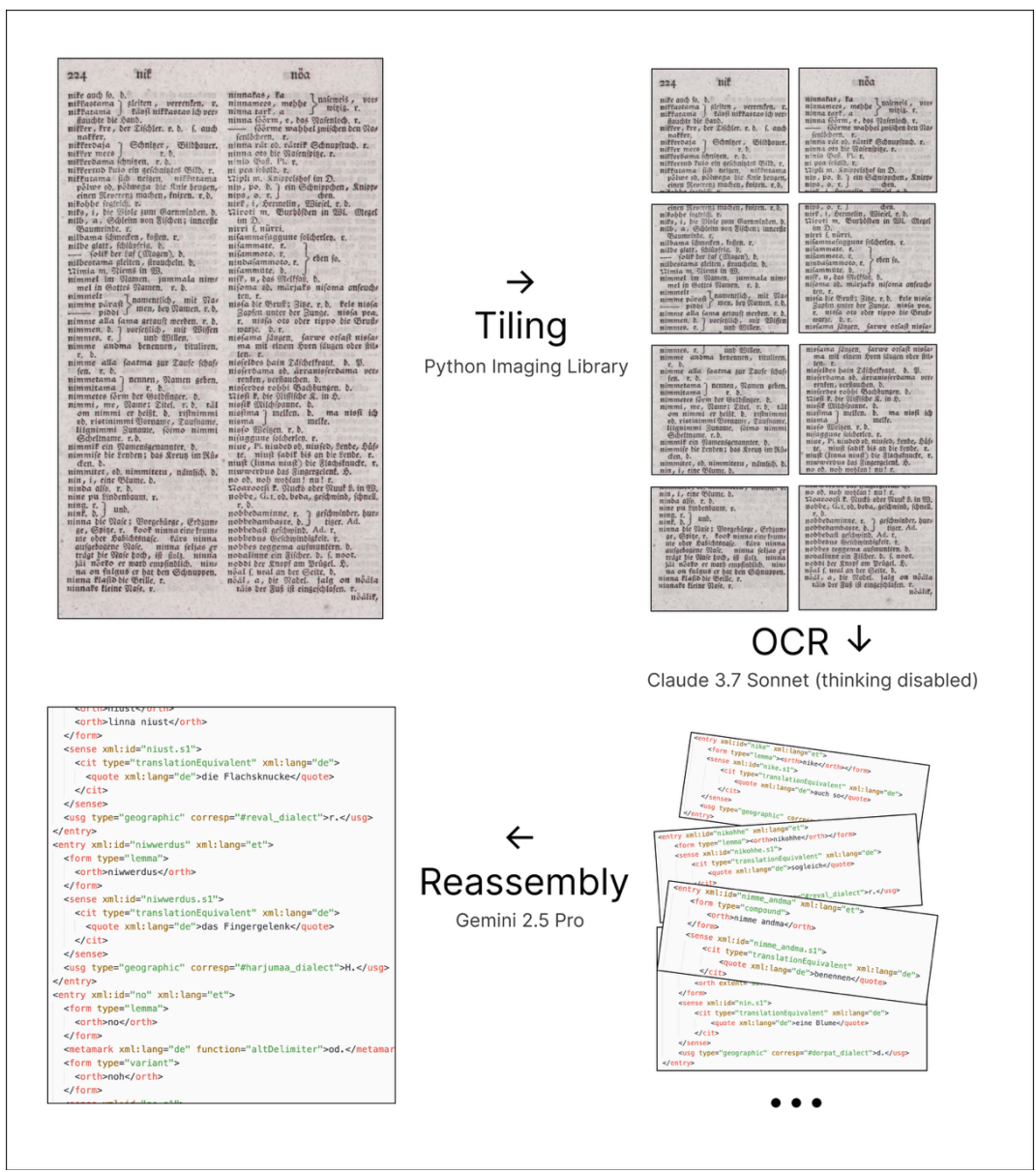

Figure 3. The tiling, OCR, and reassembly pipeline used for structured data extraction from Hupel's dictionary

Each image segment was processed through a separate query, resulting in an XML output corresponding to that specific portion of the page. Thus, 8 individual XML objects are obtained per page, which must then be assembled into a single cohesive XML document. For this task, we employed Gemini 2.5 Pro. It is important to note that the experiment's primary objective was not to conduct a comparative analysis of language models for each specific task, but rather assess the broader viability of this multi-stage processing approach.

We compared three different image processing methods:

- **Whole Page:** The model analyzed the entire page as a single input.
- **Two Columns:** The page was divided vertically into two distinct columns.
- **Eight Segments:** The page was decomposed into eight smaller, partially overlapping segments.

| Processing type | Accuracy | | Cost (USD)* | Input tokens |
|---|---|---|---|---|
| | Structural | Textual | | |
| Whole page | 0.495 | 0.507 | **0.370** | **7,184** |
| Two columns | 0.572 (+15.6%) | 0.687 (+35.5%) | 0.536 (+45%) | 13,988 (+95%) |
| 8 segments | **0.647 (+30.7%)** | **0.710 (+40.0%)** | 1.050 (+184%) | 57,186 (+696%) |

* Calculated according to the current pricing for Claude and Gemini (including both input and output tokens).

Table 4. Comparison of preprocessing methods, with parenthetical percentages indicating the increase relative to the 'Whole page' baseline.

A clear inverse correlation was observed between the size of the image fragments and the accuracy of the results. This effect was particularly pronounced in text transcription accuracy; segmenting the page into two columns improved accuracy by 35.5%, while an eight-segment division yielded a 40.0% improvement over the baseline whole-page processing method. However, this enhanced accuracy came at a significant cost. Although the eight-segment approach produced the most accurate results, its computational expense was 184% higher than that of the conventional method.

**Summary of Chapter 3**

The experiments with A. T. Helle's 1732 "Vocabularium" and A. W. Hupel's 1780 "Wörterbuch" indicate the significant potential of LLMs for digitising historical dictionaries — as long as we define *digitisation* as converting scanned image files to machine-readable and machine-editable form. Drawing on our experience of digitising the 1732 "Vocabularium", we have refined our workflow for the 1780 "Wörterbuch". Unlike the previous zero-shot approach used for the "Vocabularium", we now use a few-shot method[13] for the "Wörterbuch", in addition to the two-stage process, where the initial XML structure is generated from the image file using one model (Claude 3.7 Sonnet) and the resulting fragments are unified into a cohesive document by another model (Gemini 2.5 Pro). A potential future refinement to this workflow is the incorporation of the latest models from Transkribus or similar specialised software.

## 4. Future Plans

Our experience thus far has shown that applying LLMs in historical lexicography offers significant potential for considerable savings in time and cost. While this is less surprising for major languages like German, which are also represented in the dictionaries discussed here, the more notable finding is the models' ability to successfully recognise 300-year-old Estonian word forms and expressions. This is remarkable given that little, if any, such historical data was likely included in their training. This success is likely attributable to a combination of leveraging German contextual cues in these dictionaries and sophisticated pattern recognition between historical and contemporary Estonian forms.

As a result of this work, all primary early Estonian lexicographical sources, from Stahl (1637) to Hupel (1780), will be available to researchers in a machine-readable format. This dataset, currently existing as separate collections, will soon enable the aggregation of these individual sources in a single, unified resource. Such a "Dictionary of Dictionaries" or essentially a comprehensive cross-source database will encompass the entire content of 17th and 18th-century Estonian dictionaries published in print to date.[14] For researchers of language history, access to such a dataset would be of invaluable worth.[15]

The "modern" period of Estonian lexicography begins with F. J. Wiedemann's 1869 dictionary. Should the scope of the "Dictionary of Dictionaries" be expanded beyond early sources to encompass this modern period, including Wiedemann's work, early orthographic dictionaries, and other key milestones, the aggregated dataset would have further applications. Its word forms, for instance, could serve as the basis for an additional "historical" information layer within the Estonian "Combined Dictionary" (ÜS 2025). Such an integration would enable users of the currently most extensive Estonian general dictionary to trace the historical occurrences of contemporary vocabulary on a diachronic timeline[16].

In addition to digitising data and displaying it in a user-friendly manner, the authors plan to develop a unified cross-source dictionary query system named Wilhelm, which aims to save historical linguists' time by identifying individual words and word combinations across numerous datasets, including general and dialect dictionaries, historical dictionaries and language corpora, the Bible translation concordance (Ross et al. 2020), and the University of Tartu old written language corpus (VAKK). The goal of this software is not to simply replicate the entire contents of existing dictionaries into a single interface. Instead, it will use a language model to synthesise concise overviews of entries from the source data, providing precise source references (URLs) to enable straightforward verification of all presented information for any given linguistic unit.

---

[13] In a *few-shot* approach there is *some* example data given to the language model.
[14] This would be comparable to the dictionary of old Estonian Bible translations (Käsi et al., 2025).
[15] A similar attempt has been undertaken in the University of Tartu, although the current (Sept. 2025) version of the "Dictionary of Historical Written Estonian" (digi-VAKS), based on the initial project description for a comprehensive Estonian historical dictionaries' database (Prillop & Ress 2013), only contains a fraction of the total vocabulary scope contained in the historical dictionaries. The main principles contained in this research will be followed for the creation of a new "Dictionary of Dictionaries".
[16] Tiina Paet has undertaken the preparations for separate research to plan this new presentation of historical linguistic data within general language data dictionary entries.

**Conclusion**

In historical lexicography, LLMs readily enable the creation of necessary scripts and programmes that facilitate work with centuries-old sources and their derived datasets. The pattern-recognition capabilities of language models provide a foundation for extending the functions of traditional text recognition tools, particularly in the analysis of scanned image material.

Historical Estonian dictionaries are often opaque to contemporary readers, as over the centuries significant orthographical shifts in both Estonian and German entail substantial differences from the modern standard language. To make these dictionaries more accessible, their content can be enriched semi-automatically with the aid of language models, in addition to manually adding modern word forms and meanings. Although manual analysis is still needed for tasks like combining information and finding context, language models are capable of enriching these old dictionaries when provided with sufficient information. This, in turn, can make even the most challenging content accessible to modern users (e.g., Vestring 1998 [17XX]; Helle 1732; Hupel 1780; Hupel 1818).

Large language models also show great potential for text recognition. Since the advent of vision-enabled models (like GPT-4V in September 2023 and subsequently others, including Anthropic's Claude models), researchers have been able to process image files directly. We were able to extract text directly from image files without requiring external preprocessing steps since August 2024. This capability allowed us to convert Anton Thor Helle's 1732 dictionary into a machine-readable and machine-editable format with minimal cost and time expenditure: the language model not only performed OCR on the Fraktur script, but also structured the resulting data into JSON format, which was easily converted to a CSV table for manual editing. Building on this experience, we are now experimenting with August Wilhelm Hupel's 1780 dictionary: this time using a two-stage approach, with one language model processing page fragments to create initial XML outputs, and a second one for unifying these into a cohesive document. With this new approach, we expect to achieve a more granular, higher-quality result, thereby minising the need for manual correction. At present, retaining a human editor in the loop is unavoidable; however, considering the rapid pace of LLM development, the need for human intervention is likely to decrease significantly in the future.

All the experiments and tangible results presented in this article illustrate the potential of language models not only for historical lexicography but also for the study of dictionaries for languages with limited representation in LLMs, not comparable to the representation of major languages, primarily English, but also German. While the presence of German significantly aids the analysis concerning the (bilingual) dictionaries described here, there is reason to believe that LLMs can (at least to some extent) also understand and analyse historical Estonian even without this Germanic scaffolding (Jürviste et al. 2025a: 76–80). Although it is difficult to predict the developments of language models in the upcoming years, they already are powerful tools assisting in the analysis of historical dictionaries and processing the information thereof.

**Author Information**

**Madis Jürviste**
Institute of the Estonian Language: Lexicographer, Junior Researcher; University of Tartu: Junior Researcher. Main research topics: Estonian 17th and 18th century lexicography; studying the potential application of large language models for practical applications in analysing and creating lexicographical resources.
Roosikrantsi 6, 10119 Tallinn, Estonia
madis.jyrviste@eki.ee
ORCID iD: 0009-0003-7496-7097

**Joonatan Jakobson**
University of Tartu: BA-student (Estonian and Finno-Ugric Linguistics). Main research interests: History of Estonian literary language and Bible translation; Integrating large language models, machine learning and corpus linguistics for historical text digitisation and analysis.
Jakobi 2, 51005 Tartu, Estonia
joonatan.jakobson@ut.ee